\documentclass[]{IOS-Book-Article}

\usepackage{nesybook}

\usepackage{chapter-dummy/chap}

\begin{document}

\begin{frontmatter}              

\title{Compendium of Neuro-Symbolic Artificial Intelligence}


\author[A]{\fnms{Pascal} \snm{Hitzler}}, 
\author[B]{\fnms{Md Kamruzzaman} \snm{Sarker}},
and 
\author[A]{\fnms{Aaron} \snm{Eberhart}}

\runningauthor{Pascal Hitzler, Md Kamruzzaman Sarker, Aaron Eberhart}
\address[A]{Kansas State University}
\address[B]{Kansas State University}

\end{frontmatter}

\maketitle

\setcounter{tocdepth}{0}
\tableofcontents

\setcounter{page}{0}
\chapter{The Roles of Symbols in 
Neural-based AI: \\ They are Not What You Think!}
\label{chapter-silver-mitchell:chap}
\chapterauthor{Daniel L. Silver}{Jodrey School of Computer Science, Acadia University, Wolfville, NS, Canada B4P2R6}
\chapterauthor{Tom M. Mitchell}{Machine Learning Department, Carnegie Mellon University, Pittsburg, PA, USA, 15213}

\allchapterauthors{Daniel Silver, Tom Mitchell}

 ABSTRACT: We propose that symbols are first and foremost external communication tools used between intelligent agents that allow knowledge to be transferred in a more efficient and effective manner than having to experience the world directly. But, they are also used internally within an agent through a form of self-communication to help formulate, describe and justify subsymbolic patterns of neural activity that truly implement thinking.  Symbols, and our languages that make use of them, not only allow us to explain our thinking to others and ourselves, but also provide beneficial constraints (inductive bias) on learning about the world. 
 In this paper we present relevant insights from neuroscience and cognitive science,  about how the human brain represents \emph{symbols} and the \emph{concepts} they refer to, and how today’s artificial neural networks can do the same. 
 We then present a novel neuro-symbolic hypothesis and a plausible architecture for intelligent agents that combines subsymbolic representations for symbols and concepts for learning and reasoning. 
 Our hypothesis and associated architecture imply that symbols will remain critical to the future of intelligent systems NOT because they are the fundamental building blocks of thought, but because they are characterizations of subsymbolic processes that constitute thought. 

\section{Introduction}

One of the great open questions regarding intelligence is - How do complex neural systems such as the brain or deep neural networks successfully combine symbolic and subsymbolic representations and processing?  How is it that we humans, with brains consisting of complex networks of neurons, succeed in producing and consuming the symbolic knowledge we communicate through natural language?  Similarly, in designing intelligent agents based on artificial neural networks, what is the most effective way to integrate symbolic and subsymbolic approaches?

We suggest here that symbols and symbolic reasoning are communication tools used externally and internally within an agent to explain decisions, to help guide reasoning, and to bias learning.  Symbols are based on categories that help humans make sense of a complex world, individually and as groups.  They provide the means by which agents can learn concepts with each other more efficiently and effectively using less energy and with lower risk. But symbolic representations are of even greater importance to agents because: (1) they can also be used internally in a form of self-communication
to describe, justify, and guide subsymbolic patterns of neural activity, and (2) they  provide an inductive bias to guide future learning of new symbols, concepts and their relations.

Below we begin by defining a terminology for discussing the neural encoding of symbols and concepts, and describe the key questions we seek to answer about combined neuro-symbolic systems.  We then present relevant research results from neuroscience, behavioral (cognitive) science, and artificial intelligence, that yield evidence about the combination of symbolic and subsymbolic processing in humans and current artificial neural networks.  Guided by this evidence, we present a novel neuro-symbolic hypothesis and an associated architecture meant to provide a plausible answer to the question of how humans might implement neuro-symbolic reasoning, and how future intelligent agents might be designed to do so as well.



\section{Problem Formulation}

This section defines the terminology we will use to precisely discuss symbolic and subsymbolic representations and reasoning, and presents the questions considered in this paper.  It then presents a variety of research results from neuroscience, behavioral science, and artificial intelligence which are relevant to understanding how symbolic and subsymbolic processes may interact to produce intelligent behavior.

\subsection{Definitions}

To discuss symbols and their relationship to connectionist learning, as well as their relationship to neural activity in the human brain, we begin by defining what we mean by a concept and a symbol.

\begin{quotation}
{\bf Definition 1.} A {\bf concept} is an object, a collection of objects, or an abstract idea that can be learned and represented by an intelligent agent.
\end{quotation}

Concepts may range from specific physical objects (``that hockey puck"), to a category of objects (``birds"), to very abstract and semantically complex ideas (``blue”, ``top”, ``justice”, ``try”, ``meaning”).  More complex concepts can be built out of multiple more primitive concepts ``girl riding a bike", ``writing a technical paper").

\begin{quotation}
{\bf Definition 2.} A {\bf symbol} is a mark, sign or an object that represents, or refers to, some concept.
\end{quotation}

For example, an image of a sports mascot is a symbol that refers to a particular sports team, a written word such as ``peach” is a symbol that refers to a particular class of foods, a statue of a famous politician is a symbol that refers to a specific human being.  Many of our symbols and their meanings (the concepts to which they refer) are shared across our culture \cite{jung1968}, including symbols such as sports mascots, words, and statues.  We also have personally meaningful symbols whose meanings may not be shared across our culture, such as a souvenir purchased during a particular vacation, which may act as a symbol that to us refers to that vacation.  Words serve as symbols in natural language, and we will also treat sentences as symbols that represent more complex concepts (``girl riding a bike").
Note that with respect to C.S.Peirce a founding father of semiotics, we use the term symbol to be his equivalent of the most general class of ``signs" that includes the categories of ``icon" and ``index" and  ``symbol" \cite{CSPeirce_Wikipedia}

\begin{quotation}
{\bf Definition 3.} We will call an agent's internal neural activity that encodes a symbol its {\bf symrep} for that symbol (short for the symbol's neural representation). 
\end{quotation}

For example, when a person sees the written word ``peach", their visual cortex generates neural activity that represents this letter string symbol. We call this neural activity the symrep of the symbol ``peach".

\begin{quotation}
{\bf Definition 4.}  We will call an agent's internal neural activity that encodes the concept referred to by a symbol its {\bf conrep} (short for that concept's neural representation). 
\end{quotation}

For example, when a person views the written word ``peach," their brain generates (1) neural activity that encodes this symbol (i.e., the symrep of the letter string ``peach"), and (2) neural activity that encodes the concept this word refers to (i.e., its conrep).  
In contrast, when a person sees an actual peach, their brain generates neural activity that encodes this concept (i.e., its conrep).  Depending on the context, their brain may or may not also generate the activity that encodes the symbol for peach (i.e., the symrep of the letter string ``peach").
Below we present results from brain imaging studies that reveal some of the properties and timing of these symrep and conrep patterns of neural activity.

In this paper we are particularly interested in human and deep neural network agents, and the roles that symbols may play in their mental lives.  Toward this end we make the following assumptions:

\begin{quotation}
{\bf Assumption 1.}  Humans and AI agents based on deep neural networks encode both symbols and the concepts they refer to, using distributed vectors of neural activity.   
\end{quotation}

\begin{quotation}
{\bf Assumption 2.} It is the network's connection weights that must be learned from experience, to generate the appropriate neural activations (symrep and conrep) based on input stimuli and context.
\end{quotation}

These assumptions allow us to consider that both deep networks and the human brain share the same style of encoding: representations consisting of distributed neural activation.  Notice this is the assumption that connectionist studies of cognitive science have made for many years \cite{rumelhart1988, FELDMAN1982}.  Figure~1 illustrates this, by showing the distributed pattern of fMRI neural activity evoked by the word “airplane” in a human brain, and the distributed pattern of neural activity that encodes “airplane” in Word2Vec, a frequently used distributed representation learned by an artificial neural network \cite{Mikolov2013}.   

\begin{figure}[ht]
    \centering
    \includegraphics[width=0.97\textwidth]{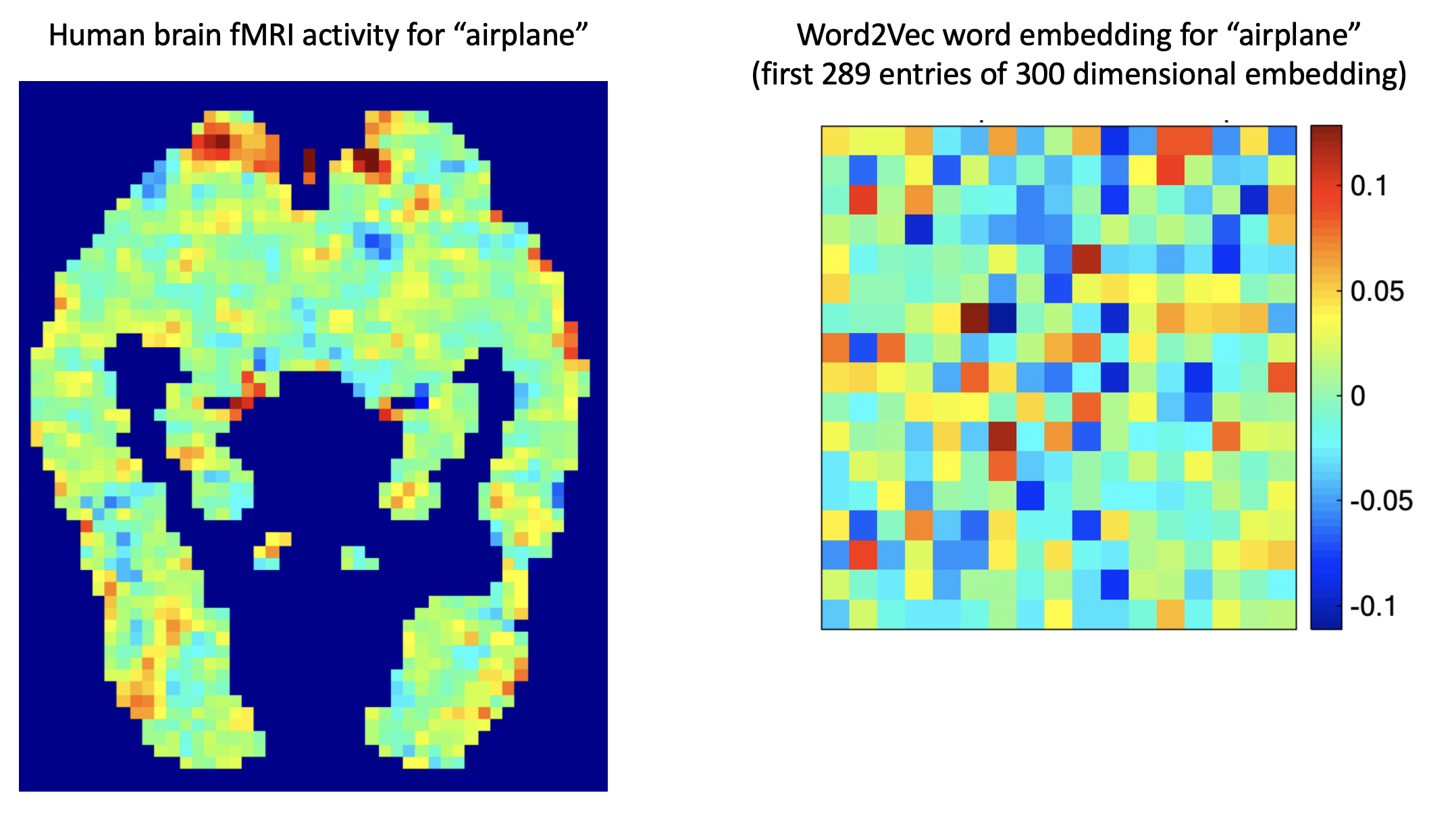}
    \caption{ {\bf Distributed encodings for the concept (conrep) referred to by the symbol “airplane.”}  Left panel shows one 2D slice of the 3D image of fMRI neural activity when a subject reads the word “airplane.” Red indicates high activation, blue indicates low (dark blue indicates regions outside cortex) (reprinted from \cite{Mitchell2008}).  Right panel shows part of the 300 dimensional distributed encoding for “airplane” learned by an artificial neural network trained to predict masked-out words in text, and available as Word2Vec \cite{Mikolov2013}.}
    \label{fig:distributedEncodings}
\end{figure}

Note that the distributed representations of concepts can be highly multimodal.  In the human brain, for example, there is strong evidence that neural activity evoked by thinking about physical objects is spread across a variety of distinct brain regions associated with different sensory and motor actions. For example, thinking about the concept refered to by the symbol “tomato” has been found to evoke neural activity in a brain region called gustatory cortex which is associated with the perception of taste, in an additional region called parietal cortex where neural activity encodes the approximate size of an object, in an additional region called pre-motor cortex, which is involved in planning of hand movements such as grasping a tomato, etc. \cite{Mitchell2008}.   Note also that the exact pattern of neural activity representing the concept might vary by context and by attention the agent places on different parts of the representation within that context (e.g., by whether the agent is considering eating or throwing the tomato).  In this paper we will treat the brain-wide pattern of neural activity observed when a concept is considered in isolation as the single distributed representation of the concept, and we will consider the context-specific variants of this pattern of activity to be the result of computations (e.g., attention and other neural processes) operating on that unique generic representation of the concept.

\subsection{Key Neuro-Symbolic Questions}

We now have a basic set of definitions and a vocabulary for discussing the main questions addressed in this paper.  We are particularly interested in the role of symbols in the mental life of human and deep neural network agents.  The following are some of the key open questions in  neuro-symbolic research:

\begin{enumerate}

    \item Given the use of symbols for communication between agents, how do agents map between symbolic language and subsymbolic processes?
    
    \item Do symbols play an internal role for the agent, beyond external communication? For example are symbols potentially useful in the agent’s reasoning processes, in learning, in memory storage and retrieval?
    
    \item Given the human brain's obvious ability to manipulate symbols, how does reasoning combine symbol-level and subsymbolic representations and processes?  For example: young children, even before they have shared cultural labels for objects, can identify and express interest in objects (eg. mother, juice).  Does this mean symbols are not required for reasoning?
    
    \item Context plays an important semantic role in understanding any sequence of percepts and the decisions and actions that are taken based on those percepts. How is context formulated and used? 
    
    \item Attention to the most relevant things is required by an agent in order to survive in a world using a finite computing system. What are the sources of attention?  
    
    \item How do we integrate/consolidate new concepts and symbols into our memories?  Are there limits to such memory?

    \item How do we imagine new solutions to challenging complex problems? Is this done by manipulating symbolic representation (symrep),  concept representation (conrep) or both?

    \item How and why did our brains develop to manipulate both symbolic and non-symbolic representations?
    
\end{enumerate}  

Although all of these issues are interesting and important, in the remainder of this paper, we focus on the first three of these questions.

\section{Background}

Before considering our hypotheses about how biological and computer agents do or should use and represent symbols, it is useful to first review what is currently known about how the human brain represents symbols and the concepts they refer to, and how today's artificial neural networks do so as well.  The following subsection summarizes some current results obtained from brain imaging studies of humans.  The ensuing subsection provides an overview of a recent cognitive model that has drawn much attention and is closely related to our theory.
The final subsection considers some of the known properties of the distributed representations learned by artificial deep neural networks, highlighting commonalities and differences.

\subsection{Concepts, Symbols and Their Representation in The Brain}

The advent of fMRI in the mid-1990’s made it possible for the first time to noninvasively observe correlates of neural activity in the brain with a spatial resolution of a few millimeters and a temporal resolution of approximately one second.  This has led to many experimental studies of distributed neural activity associated with various symbols and the concepts to which they refer. 
Furthermore, other brain imaging technologies such as electroencyphalography (EEG) and magneto-encephalography (MEG) allow higher-speed imaging, providing up to one image per millisecond, with lower spatial resolution (e.g., approximately 1-2 cm for MEG) \cite{Sudre2012}. Below we summarize some of what has been learned about the representation of symbols and concepts in the human brain.

\begin{figure}[ht]
    \centering
    \includegraphics[width=0.8\textwidth]{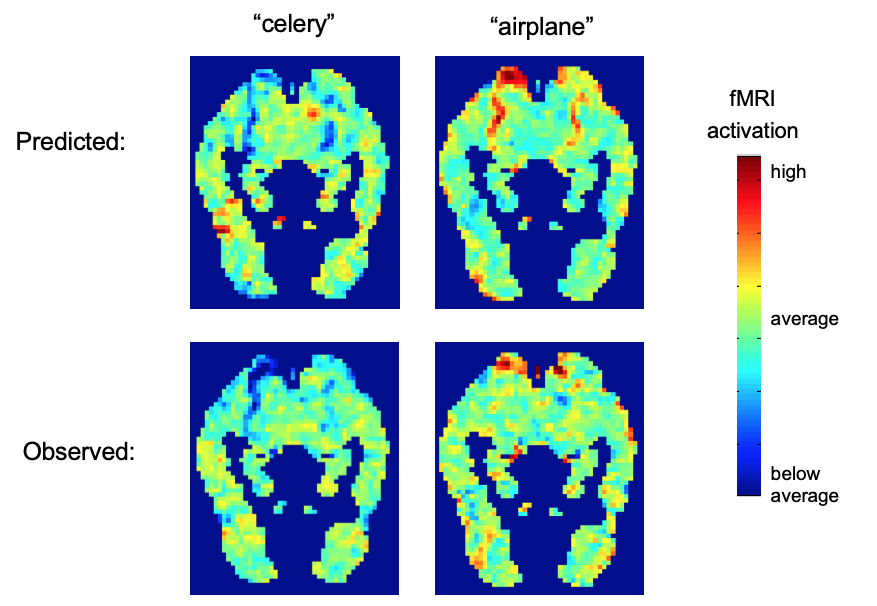} 
    \caption{{\bf Predicted and observed fMRI images for ``celery" and ``airplane" after training on other nouns.} Top row: the predicted fMRI images for stimulus words ``celery" and for ``airplane", by the trained model of \cite{Mitchell2008} which observed neither of these words during training.  Bottom row: the observed fMRI images when the human subject read these two words. Though the predictions are imperfect, they capture significant features of the observed data, implying a systematicity in the patterns of neural activity associated with noun meanings in the brain.}
    \label{fig:brainPredictedAndObserved}
\end{figure}

\begin{itemize}
    \item 
{\em Reading of isolated nouns leads to repeatable, distributed patterns of neural activity.}  For example, the words “hammer,” “house” and “airplane” each produce a  distinct, repeatable pattern of neural activity, distinct enough that it is possible to train a machine learning classifier to distinguish which of these words is being read, based on observed fMRI activity of a human subject while they read each of these words in isolation \cite{Mitchell2004, shinkareva2008}, 
or while they view photos of these concepts \cite{haxby2001}.  For example, Figure \ref{fig:distributedEncodings} shows one 2-dimensional slice of the 3-dimensional image of fMRI activity while one person read the symbol ``airplane."

\item
{\em These patterns of neural activity differ from word to word, and words with similar meanings exhibit similar patterns of activity.}  For example, the observed distributed patterns of neural activity for the three words “celery,” “tomato,” and “airplane” differ, but the pattern for “celery” is more similar to that of “tomato” than “airplane”, as measured by cosine similarity between the patterns.  In fact, the cosine similarity between neural representations  of two words in the brain correlates highly with the cosine similarity between their Word2Vec representations used in deep  neural network language models.

\item
{\em The brain's neural activity for any given concept is very similar regardless of whether the stimulus symbol is an English word, Portuguese word, or a picture.}  This suggests the final neural activity generated in response to the stimulus symbol is dominated by the full representation of the concept (the {\em conrep}), with relatively less neural activity allocated to representing of the symbol itself (its {\em symrep}) \cite{Just2012}.

\item
{\em These neural activity patterns per concept are highly similar across different human subjects.}  In fact, the neural activity pattern that forms the conrep of a concept is so similar across people that it is possible to decode which word a {\em new person} is reading, based on their fMRI or MEG neural activity, using a machine learning classifier trained on brain image data from {\em other people} \cite{shinkareva2008}.

\item
{\em There is a strong systematicity in the neural activity that represents different words' meanings (their conreps)}, such that a 25 dimensional latent representation suffices to predict the observed 10,000 dimensional voxel neural activity patterns of concrete nouns quite well \cite{Mitchell2008}.  \tom{clarify} Put another way, the $10^4$ dimensional patterns of fMRI neural activity associated with different nouns are not independent random neural activity patterns, but appear to lie in a much lower dimensional space.  Evidence for this includes the fact that it is possible to train a machine learning model to {\em predict} the fMRI activity that will be observed for new concrete nouns that it was not trained on, based on corpus statistics of the new noun.  For example, in \cite{Mitchell2008} we present a trained model that predicts the fMRI activation for new nouns well enough that it achieves 79\% accuracy in deciding which of two new nouns outside those it was trained on were associated which which of two new fMRI images it had never seen.  Here, each noun was represented by a vector of frequencies with which it co-occurred with each of 25 verbs such as ``eat" and ``run" within a trillion word text corpus.   Figure \ref{fig:brainPredictedAndObserved} shows the predicted and observed fMRI images for two typical nouns.  
\end{itemize}

To summarize, the above fMRI brain imaging studies reveal important properties of the spatially distributed patterns of neural activity used by the human brain to encode stimulus symbols and the concepts they refer to.  These neural activity patterns include both the symreps of the stimulus symbols and the conreps of their associated concepts, but they are dominated by the conreps (we see nearly the same activity whether we present the English symbol ``house" or the Portuguese equivalent symbol ``casa"). Furthermore, they are widely distributed across the brain (especially in sensory and motor regions associated with the concept referred to) \cite{PULVERMULLER2013}, they are suprisingly similar across individuals (so much so that a classifier trained on one person can successfully decode the noun a new person is reading based on their brain activity), and their neural activity encodings have a strong systematicity that allows predicting the fMRI image that will be seen if a new noun is presented to the person.

One property of these neural activation patterns that is not revealed by fMRI is the {\em sequence} of neural activity that occurs as a person perceives a word symbol, and as the brain creates the conrep for that symbol. This sequential process of word reading is not easy to study with fMRI because it is a subsecond process for the brain, and fMRI supports only about one image per second. 

Fortunately, a second brain imaging technology called magnetoencephalography (MEG) allows capturing whole brain images every millisecond with effective spatial resolution of about 1.5 cm. By MEG imaging people while they read words we can study questions such as ``what is the sequence of neurally encoded information about the symbol (its symrep) and the concept it refers to (its conrep) that appears at different locations and at different times in the brain during word reading?". For instance, does the brain first perceive the symbol, and then activate the entire encoding of the associated concept at some single point in time, or does the concept's neural encoding appear gradually, first one part and later another?  Below we summarize some of what has been learned about this question.  Much of these results are from a study in which human subjects in a MEG scanner viewed stimuli consisting of a concrete noun and a line drawing of the concept referred to by  the noun \cite{Sudre2012}, and from subsequent analyses of this experimental data.  More recent experiments have also reported results consistent with these findings \cite{Sassenhagen2020}.   

\begin{figure}[ht]
    \centering
    \includegraphics[width=0.99\textwidth]{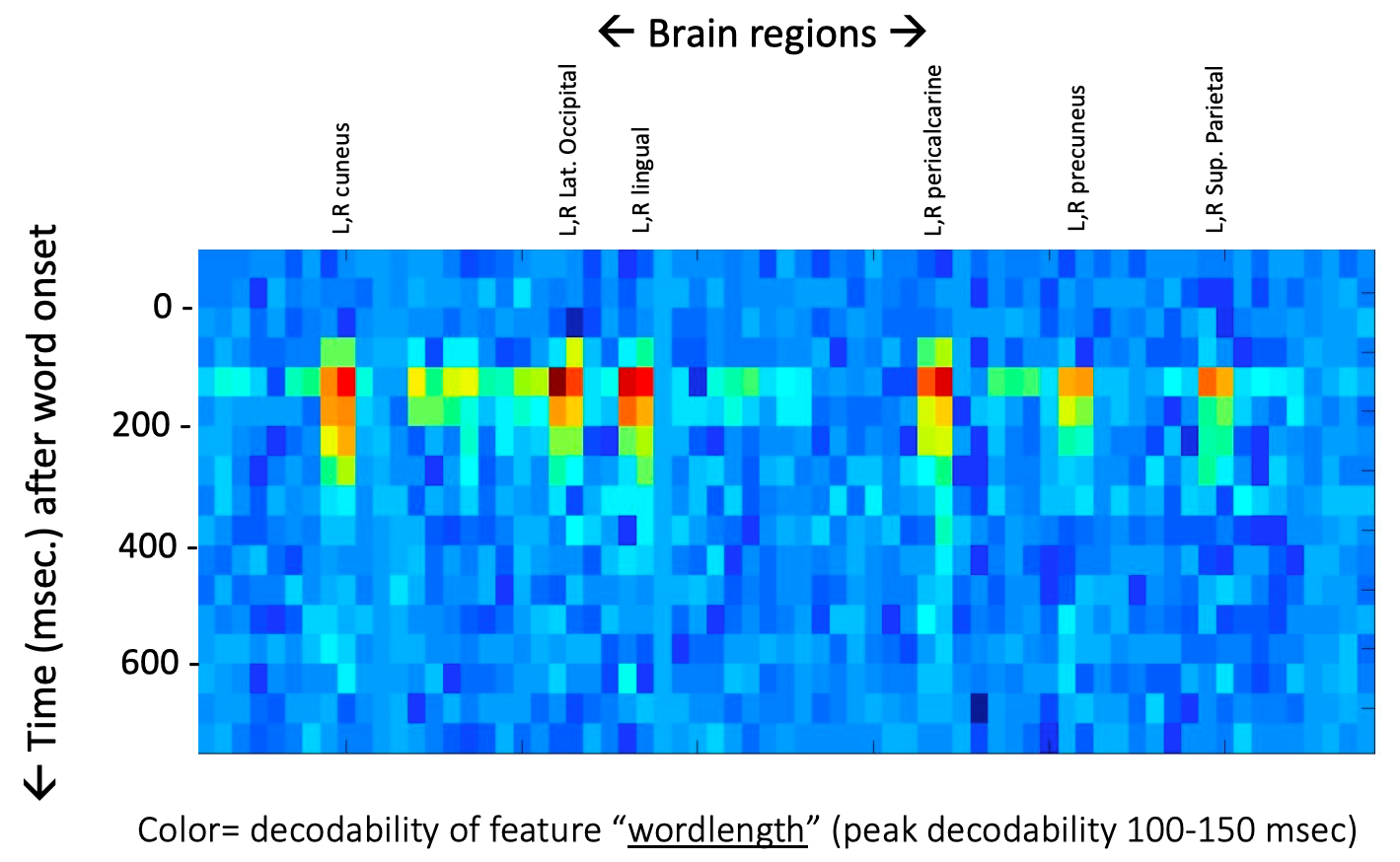} 
    \caption{{\bf Decodability of a word symbol feature (the number of characters in the stimulus word) from MEG data, over time.} The vertical axis shows time in milliseconds since stimulus word onset, and the horizontal axis indicates different brain regions.  The redness of each point shows how well the number of characters in the word can predict the MEG activity in the corresponding brain region, during the corresponding 50 msec window of time (percent of MEG variance predicted, averaged across nine human subjects).  Note this perceptual feature of the stimulus word is decodable beginning at 100-150 msec after stimulus onset, but the encoding of this feature disappears by 300 msec post stimulus onset.  This figure is based on analysis of the data from \cite{Sudre2012}.}
    \label{fig:decodabilityOfWordLength}
\end{figure}

\begin{figure}[ht]
    \centering
    \includegraphics[width=0.99\textwidth]{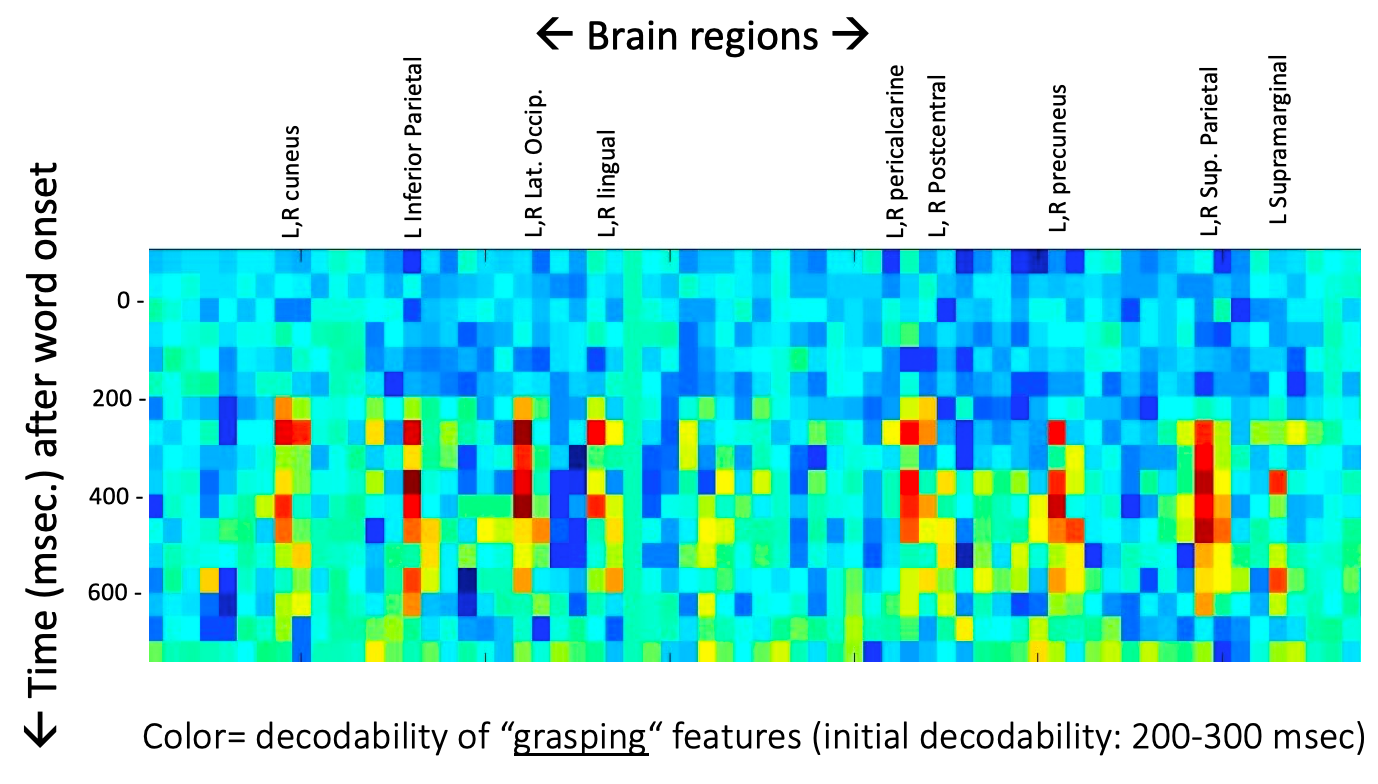} 
    \caption{{\bf Decodability of word semantic feature from MEG data, over time.}  The decodability of whether the stimlus word describes an object that is ``graspable," in different brain regions and at different 50 msec windows of time after stimulus onset.  Note this plot follows the same format as the plot in Figure \ref{fig:decodabilityOfWordLength}, but here we consider a semantic feature (graspability) of the conrep associated with the stimulus, whereas the previous figure considers the perceptual feature (word length) of the stimulus symrep. Note this semantic feature of the concept is decodable beginning around 200-250 msec, and persists through 500 msec after stimulus onset -- much longer than the perceptual word length feature of Figure \ref{fig:decodabilityOfWordLength}.  This figure is based on analysis of the data from \cite{Sudre2012}.}
    \label{fig:decodabilityOfGraspability}
\end{figure}

\begin{itemize}
\item
{\em It takes approximately 400 msec for the brain to perceive a symbol and activate the associated conrep.} This is a widely accepted result in neurolinguistics, as it is supported by multiple types of evidence.  First, if we train a machine learning classifier to decode the word being read from MEG brain activity, and use only a 50 msec subwindow of the MEG data, then we find that the 50 msec subwindow at which this decoding is most accurate is centered at 400 msec post stimulus onset.  Second, other researchers have observed using EEG imaging that if a person is given a highly surprising word in a sentence (e.g. ``I put the milk in the yellow") then a large neural surprise signal is visible at 400 msec after the appearance of the surprising word, suggesting this is the point at which the brain first detects that the concept being referred to by the word symbol has an unexpected meaning.

\item
{\em The neural encoding of the word symbol itself (the symrep) appears at roughly 100-150 msec post word stimulus onset.} In order to determine this, one can ask the question ``at what point in time is it possible to decode from neural activity the perceptual properties of the stimulus symbol that are independent of the concept to which it refers?"  One such perceptual property is the number of characters in the stimulus word (e.g., does the word have greater than 4 characters or not?).  We find that this property can be reliably decoded (i.e., with 80-90\% accuracy) from a 50 msec window of MEG neural activity between 100-150 msec after stimulus onset, but no sooner than that.
Figure \ref{fig:decodabilityOfWordLength} shows the times after stimulus word onset that the word length can be decoded in different brain regions.  Although the full symrep might include additional neural activity not detectable by MEG imaging, this shows that at least by this point in time the brain activity does represent aspects of the symbol appearance.  As shown in this figure, at the time when word length is first decodable from observed MEG actiivty, it is decodable independently in each of half a dozen distinct brain regions in visual and parietal cortex, though little is known about why.  Also, by 300 msec post stimulus onset, the information about word length can no longer be decoded from neural activity, suggesting this part of the symrep may be a transient used only as part of a larger process to index into the word's associated conrep.  

\item
{\em The neural encoding of the concept associated with the stimulus symbol (the conrep) appears gradually and cumulatively over time, not instantaneously, beginning at approximately 150-200 msec post stimulus onset, and continuing through 450 msec}. The evidence for this is that different semantic features of the conrep  become decodable from neural activity at different times.  For example, one of the earliest decodable semantic features is animacy (whether the word describes a living thing) which is first decodable from neural activity at approximately 150-200 msec post stimulus onset.  In contrast, the semantic feature of whether the stimulus word describes something ``graspable" is first decodable at 200-250 msec post stimulus onset.  Figure \ref{fig:decodabilityOfGraspability} displays the decodability of the ``graspable" feature of the stimulus word, over time, in different brain regions. Similar to the symrep, once these semantic features are decodable, we find they are simultaneously decodable independently from a number of disparate brain regions. However, unlike the symrep, these semantic features decodable from the emerging conrep persist (continue to be encoded by neural activity) through at least 500 msec.  These results using MEG imaging are corroborated by an independent study using EEG imaging which reports that significant aspects of EEG neural activity can be predicted from word semantics during 300-500 msec post word onset \cite{Sassenhagen2020}.

\item 
\emph{The human brain recognizes objects in images approximately as quickly as it recognizes letter strings as words.}
It has been shown that humans can recognize an animal in an image after as little as 150 msec, and that this reaction time does not statistically improve with practice \cite{FabreThorpe2001ALT}.
Note this is roughly the same amount of time it takes for neural activity to reflect perceptual features of an observed word, as summarized in Figure \ref{fig:decodabilityOfWordLength}.

\item {\em The two types of human visual attention exhibit timing similar to symbol recognition and concept representation.}  Studies of visual attention in humans reveal that there are two distinct types of visual attention.  {\em Exogenous attention} is stimulus-driven and involuntary, and typically occurs at roughly 100 msec post stimulus onset.  {\em Endogenous attention} is more goal-driven and voluntary, and typically begins are roughly 300 msec post stimulus onset.  Note these 100 msec and 300 msec timings align roughly with the timings for recognizing a word (generation of the symrep), and the formation of the concept representation (generation of the conrep).  Combined with the points above, this suggests the possibility that word recognition (symrep) is a bottom up, involuntary process in contrast with the more complex representation of word meaning (conrep) which may involve more top down and context-driven processes \cite{Jigo2020, Dugu2018, Jigo2021}.

\item {\em Understanding metaphores and idioms appears to require a blend of conceptual and symbolic reasoning.} Idioms, such as ``it's raining cats and dogs", are phrases that have a meaning not deducible from those of the individual words.  Instead, this phrase requires interpretation that considers both the symbols, the the underlying concepts and things to which those concepts may be related.  Pulvermuller suggests \cite{PULVERMULLER2013} that ``compositional semantic processing of action-related words and semantic processing of abstract idiomatic constructions as a whole simultaneously and jointly contribute to idiom comprehension."  If you are asked to ``spill the beans" you understand that you need to provide some additional information and not dump out a can of beans.  This suggests that symbolic processing and subsymbolic processing (often involving the senses) need to interact during many  aspects of reasoning. 


\end{itemize}

Taken together, these results shed some light on the timing of brain processes that perceive an external symbol via the activations of a symrep and use this to create the conrep of the related concept.  If we take the neural encoding of the symbol's perceptual appearance (e.g., number of characters in the word) to be part of its symrep, then this first appears at roughly 100-150 msec post stimulus onset.  In contrast, the conrep does not fully appear until 400 msec, though aspects of the conrep (some of its semantic features) begin to appear as early as 200 msec post stimulus onset.  Given these timings, we might estimate that if the brain has an indexing process that activates the conrep from some internal symrep, then this process is a gradual one that takes about 250 to 300 msec to complete (i.e., the 400 msec post onset completion, minus the 100-150 msec appearance of the symrep).

\subsection{Symbols, Concepts and Cognitive Models}

There are many patterns of brain activity that occur about which we are not consciously aware, such as the brain activity that controls our breathing, or the neural activity that manages our balance while standing or walking. In contrast, whenever we speak words we are using symbols that refer to concepts about which we are consciously aware. That is, generating language is an intentional and conscious activity.  Similarly, when we “speak silently to ourselves” while thinking, we are also aware of the symbols and concepts about which we are thinking.  In this paper, when we use the term symbol, we are typically referring to mental activities where the human subject is aware of the patterns of brain activity. 
This is related to Daniel Kahneman's 2011 book entitled ``Thinking Fast and Slow" \cite{kahneman2011thinking} (based on earlier work by Stanovich and West \cite{Stanovich2000})  which describes two modes of human thought - Systen 1 is fast, instinctive and emotional; whereas System 2 is slower, more deliberate and more logical.

More specifically, \emph{fast} System 1 thinking is characterized as unconscious reasoning; where judgments are based on intuition and information is processed quickly; typically involving large quantities of data; effortless and automatic; influenced largely by experiences, emotions and memories; 
includes recognition, perception, spatial orientation; and
can be overridden by System 2 thinking.  System 1 thinking makes up the vast majority of human thinking, is observed in many animals, and is more primal to survival.  System 1 characteristics are thought to have developed  early in the evolution of animals \cite{Kaas2013}.  
Examples in order of complexity of things System 1 can do are:
determine the distance between two objects;
localize the source of a specific sound;
display disgust when seeing a gruesome image;
drive a car on a well known road; and 
understand simple phrases.

In comparison, \emph{slow }System 2 thinking is characterized by conscious reasoning; where judgments are based on controlled processes and careful examination of the precepts make for slow information flow; typically involving small quantities of data; is intentional and requires effort and control, is influenced by rules, logic and evidence; and overrides System 1 when System 1 fails to form an acceptable conclusion/decision.  System 2 thinking is predominately found in humans and other high-order animals such as the great apes, and is theorized to have developed later as a consequence of evolutionary pressure \cite{Kaas2013}.
Examples in order of complexity of things System 2 can do are:
direct your attention towards someone at a loud party;
find Waldo in a picture; 
recognize a bird call;
sustain a faster-than-normal walking rate;
count the number of A's on a page of text;
park into a tight parking space; and
solve the math problem 27 × 39 or prove the validity of the logical statement $(A \Rightarrow B) \land (B \Rightarrow C) \Rightarrow (A \Rightarrow C)$.

A key insight from Kahneman's theory is that most of the decisions made by humans are made automatically by System 1 where there is no deliberate reasoning taking place in the classical symbolic sense, whereas only a small fraction of decisions are made by System 2 \cite{kahneman2011thinking}. We like to feel that we are in control and act rationally at all times; however, Kahneman asserts that we are not really aware of much that we do, and it is only when we are asked to explain our actions that we try to explain them with ``reasons" which is often not what actually caused us to take the actions.


\subsection{Symbols, Concepts and Artificial Neural Networks}

Over the past decade, deep artificial neural networks (ANNs) have become a dominant approach to building many artificial intelligence systems.  Their use has dramatically advanced the state of the art in core AI problems such as speech recognition, computer vision, and machine translation.  Since 2018, there has been remarkable progress specifically in building very Large Language Models (LLM's) such as BERT, GPT-3, PaLM and others \cite{bommasani2021,chowdhery2022,devlin2018,GPT3_2020,Vaswani2017}.  These deep neural networks are trained to predict the next word in a sentence, the next sentence in a passage, and more.
They are trained on extremely large sets of text (e.g., many millions of pages from the world wide web) and in some cases contain hundreds of billions of trained parameters.  They have been found to have surprising abilities to generate paragraphs of human-like text, and in many cases to answer natural language questions correctly, and to explain the reasons for these answers.  

The structure and properties of these deep neural network language models are interesting in the context of this paper, because they constitute a working example of a system that successfully processes symbols (words), yet does this while internally representing the words and phrase meanings using distributed vectors of neural activations which are generated and manipulated by the network.  Furthermore, one important current research question is how to combine the knowledge these networks learn (captured in the network of interconnected artificial neuron units, and the learned strengths of their interconnections), with knowledge represented symbolically (e.g., in a database, which can symbolically represent relations such as CapitalOf(Canada, Ottawa), as well as inference rules such as ``IF CapitalOf(x,y) THEN LocatedIn(y,x)").

Some properties of these large language models that are relevant to our discussion include the following

\begin{itemize}
\item  
{\em  LLM's represent the conrep of concepts associated with word and sentence symbols using vectors of neural activation.}. Given some task that the ANN is intended to perform, the network is trained using gradient descent to tune the network parameters to maximize the accuracy of the network output prediction for each training input. The knowledge in this trained network is therefore captured in the graph of interconnections among units, and in these learned network parameters that determine the strength of interactions between individual units.  Once trained, and the network is given a new input, it calculates its output by propagating activity forward from the input, through its units, effectively creating a vector of activation at each internal network layer.  When the inputs are words, as in LLM's or translation systems, these learned parameters entail a specific activation vector for each word in the vocabulary. These activation vectors are often called {\em word embeddings} in the research literature.  Several pretrained dictionaries of word embeddings are widely used and freely downloadable on the web. For example, Figure \ref{fig:distributedEncodings} illustrates part of the Word2Vec \cite{Mikolov_Word2Vec_2013} 
embedding for the word ``airplane."

\item
{\em Word embeddings derived by ANNs trained on text from the web successfully capture much of a word's meaning, and can be used to predict neural activation of individual words in the human brain}.  Learned word embeddings such as Word2Vec have many of the same properties found in the patterns of neural activity that constitute conreps in the brain: words with similar semantics have similar word embeddings.  Interestingly, Word2Vec embeddings have been found to capture semantics in such a way that subtracting the word embedding vector for ``man" from the vector for ``king" and then adding the vector for ``woman" results in a good approximation to the vector for ``queen," showing that these learned vectors capture semantics in a way that can support some kinds of reasoning.  Furthermore, there is a close correspondence between conrep embeddings derived from ANNs and the conreps found in the human brain: word embeddings such as Word2Vec \cite{Mikolov_Word2Vec_2013} and GloVe \cite{pennington2014glove} have been used as input to machine learned models that successfully predict aspects of brain activity in fMRI and EEG data \cite{Sassenhagen2020}. This indicates that the semantics of word referents derived from large collections of web text align with the semantics reflected in neural activity in the human brain.

\item
{\em Many ANN LLMs learn to modify the context-free conreps associated with individual words, by taking into account the specific sentence context in which the word appears.}  Whereas word embedding dictionaries such as Word2Vec and GloVe capture much of the general, context independent meaning of words, they do not capture the full details of word meanings in the context of specific sentences.  For example, consider the different meanings of the word ``bank" in the sentences ``I deposited money in the bank." and ``I fished from the bank."  More recent ANNs for language processing such as ELMO, BERT, and GPT3 begin with a dictionary of context free word (and subword) embeddings, then the network modifies each word embedding based on the remainder of the sentence.  This process has been found to significantly improve overall performance of language processing ANNs across a broad range of tasks ranging from part of speech tagging, to information extraction, to machine translation.

\item
{\em In order to process each input word and generate an associated conrep, ANN LLMs learn which other words in the input text it must attend to.}
The most successful LLMs as of 2022 employ a neural network topology called a {\em transformer} architecture.  This is an autoregressive model of word sequences that employs a mechanism referred to as ``attention" \cite{Vaswani2017} which is trained to automatically determine (1) which other words in a text passage are most relevant to modifying the conrep associated with the current word, and (2) how to modify that conrep (in most architectures the modification consists of adding a learned vector to the current word's conrep). Although this mechanism does learn which other words to attend to when computing or modifying the representation of the current word, and similar processes must also be occurring in the brain, the specific computational technique called ``attention" is not necessarily the same as attention in the human brain.  

\item
{\em Intermediate representations of sentences in ANN language models have been found to align to some degree with human brain activity during sentence reading.}  For example, \cite{Jat2019} describes a MEG experiment in which human subjects read simple sentences such as ``The dog found the bone." and ``The bone was found by the dog." with words presented sequentially, one word every 500 msec. The same sentences were given to the ANN language model BERT, and a machine learning model was trained to predict the brain neural activity every 500 msec from the intermediate activations of BERT at the same point in the sentence. Predictions were found to be accurate enough to distinguish which of two new sentences a person was reading, given two new observed MEG sequences and the BERT predicted MEG activity for those sentences, with an accuracy of 90\%.  Interestingly, it was also found to be possible to distinguish earlier words in the sentence based on the BERT-predicted difference in brain activity later in the sentence (e.g., determine whether the sentence was ``The dog ate the..." or ``The girl ate the..." using only the  MEG brain activity observed during the fourth word of the sentence).  This result suggests an alignment between the learned internal representations of sentence-processing ANN language models, and processing in the human brain, which may provide a basis for a new paradigm to study and model human language processing.

\item 
{\em ANNs can be trained to map between symbol representations (symreps) and representations of their corresponding concepts (conreps), and to map from partially perceived concepts to their full representation.}  For example, \cite{DBLP:conf/flairs/IqbalS16} provides an example of such a trainable ANN system. Specifically, it describes the unsupervised training of a generative neural network (based on a stacked Restricted Boltzman Machine (RBM) architecture) that can take inputs from any of the following I/O channels: an image of a hand written digit, the spoken word saying the digit, or a robot command sequence for drawing a digit, and produce the other types as output.  For example, given just the spoken digit ``two", it can activate the top of its RBM stack which in-turn activates a large shared layer of the architecture that can represent the concept two (i.e. two's conrep) as well as other digits. The conrep, in turn, activates the other generative channels such as the representation of the written ``two." or a special symrep channel that represents each numeric digit as a one-hot 10 bit string. The paper shows that the one-hot channel can, given the appropriate symrep activation,  cause the large shared layer to create the two's conrep in the same way as any of the I/O channels. 
Thus, current ANNs already exhibit the ability to complete conreps, and to map back and forth between conreps and symreps.  



\end{itemize}



\subsection{Common ground for symbolic and neural representations}

There is much common ground across the studies mentioned above of human neural activity, human cognitive behavior, and modern artificial neural networks.  The following summarizes these shared findings: 

\begin{enumerate}

\item \emph{Consistent encoding of symbols.} 
Word (symbol) reading leads to repeatable distributed patterns of neural activity.  
There is a correspondence between ANN Word2Vec representations and representations in the human brain.  Brain representations can be predicted as a linear transformation from Word2Vec \cite{Sassenhagen2020}

\item  \emph{Encodings focus on the concept.}
Patterns of neural activity associated with symbol stimuli describe its associated concept (conrep) and  not just the symbol (symrep). In fact the encoding of a symbol such as ``cat" can be considered part of the encoding of the concept cat which also includes the sound of the word ``cat", likely prototypical images of cat, and the sense of touching a cat \cite{Volk2020}.

\item \emph{Encodings are multimodal.}
Representations in the brain and in artificial neural networks get similar patterns whether hearing or reading, and whether the word is given in Portuguese or English. The full representation is spread across different senory and motor modalities.
Evidence is that we get similar patterns whether hearing or reading, and whether the word is given in Portuguese or English.

There is also a correspondence at the level of processing multiword sentences.  Brain neural activation while reading simple sentences (e.g., The bone was found by the dog.) can be well predicted from hidden layers of BERT processing the same sentence, especially from intermediate layers \cite{Jat2019}.

\item \emph{Systematicity.}
There is a systematicity in the observed patterns of neural activity that encode word meanings in the brain, as they can predicted surprisingly well from just a 25 dimensional vector representation (embedding) or the word \cite{Mitchell2008}. Still, higher dimensional representations (e.g., 300 dimensional Word2Vec vectors) produce even more accurate predictions of observed brain activity.  Interestingly, recent transformer based ANNs such as GPT3 have shown increasingly accurate performance by increasing the size (dimension) of their embeddings. The Davinci-003 version of GPT3, released in November 2022, uses embeddings of size 12,228.  Interestingly, this is in the same ballpark as the 10,000 to 50,000  dimensional fMRI brain images achievable with modern fMRI imaging.

\item \emph{Timing. }
If the word ``cat" is the symbol for the concept cat, it takes only 100-150 msec for neural activity to encode the symbol (symrep), as evidenced by the ability to decode word length from neural activity at this point in time. However, it takes 400 msec for the brain to fully encode the concept cat (conrep) after seeing the word ``cat". This suggests it takes 250 msec to index from symrep to the conrep. 
Note activity associated with the conrep persists for hundreds of msec, unlike activity associated with the symbol appearance (symrep), which is short-lived.

\item \emph{Dual Architecture.}
Kahneman's theory of System 1 and System 2 thinking suggests that the brain learns to very quickly activate a neural pattern  Y if it is frequently and saliently coactivated with neural pattern X \cite{kahneman2011thinking}.  Hebbian learning also suggests this. So, even though the image of a peach or the symbol for a peach (symrep) is partial or vague (e.g., a smashed peach, or ``canned peaches"), the brain seems to generate the conrep for a “typical” peach in its full complexity, modulated by any available grounded perception and relevant background knowledge.

\item \emph{Sequence to sequence mapping.}
Human brains and most recurrent ANN models  are able to map sequences of symbolic inputs to sequences of outputs.  The most recent transformer ANNs use their attention mechanism (Query,Key,Value triples) to learn semantic representations that depend not only context-independent encodings of word symbols (e.g., Word2Vec), but also the sentence and paragraph contexts in which these words appear.

\end{enumerate}

From the above common links between the brain and artificial neural net studies, a picture starts to emerge. 
Much of reasoning presumably works on  distributed representations of concepts (conreps) and probably not on the symbols directly (symrep). 
But our ability to communicate means we have a way to translate back from the concepts to the symbols that refer to them and \emph{vice versa}.


\section{Our Hypotheses}
\label{hypotheses}

In this section, we propose that something very special happens when you put together (1) a system that learns to sense and represent the world and act appropriately and (2) a system that learns to sense its own internal representations (neuron activations) and relate them to shared symbols.  Most importantly, over time, these two systems work symbiotically to beneficially constrain each other during learning and reasoning.
That is, symbolic representations and symbolic processing is constrained by subsymbolic representations and processes, and vice versa. 

This section proposes a new architecture for an intelligent agent that connects neural representations to symbols. 

\subsection{A Neuro-Symbolic Hypothesis}

Our hypothesis about the relationship between symbols and subsymbolic neural representations is: 

\emph{Symbols are critical to intelligence NOT because they are the building blocks of thought, but because they are characterizations of thought that (1) allow us to explain our subsymbolic thinking to ourselves and others and (2) act as constraints on inference and learning about the world. 
Symbols explain our thinking and aid our thinking, but are not the foundation of our thinking.}

\begin{figure}[t]
    \centering
    \includegraphics[width=1\textwidth]{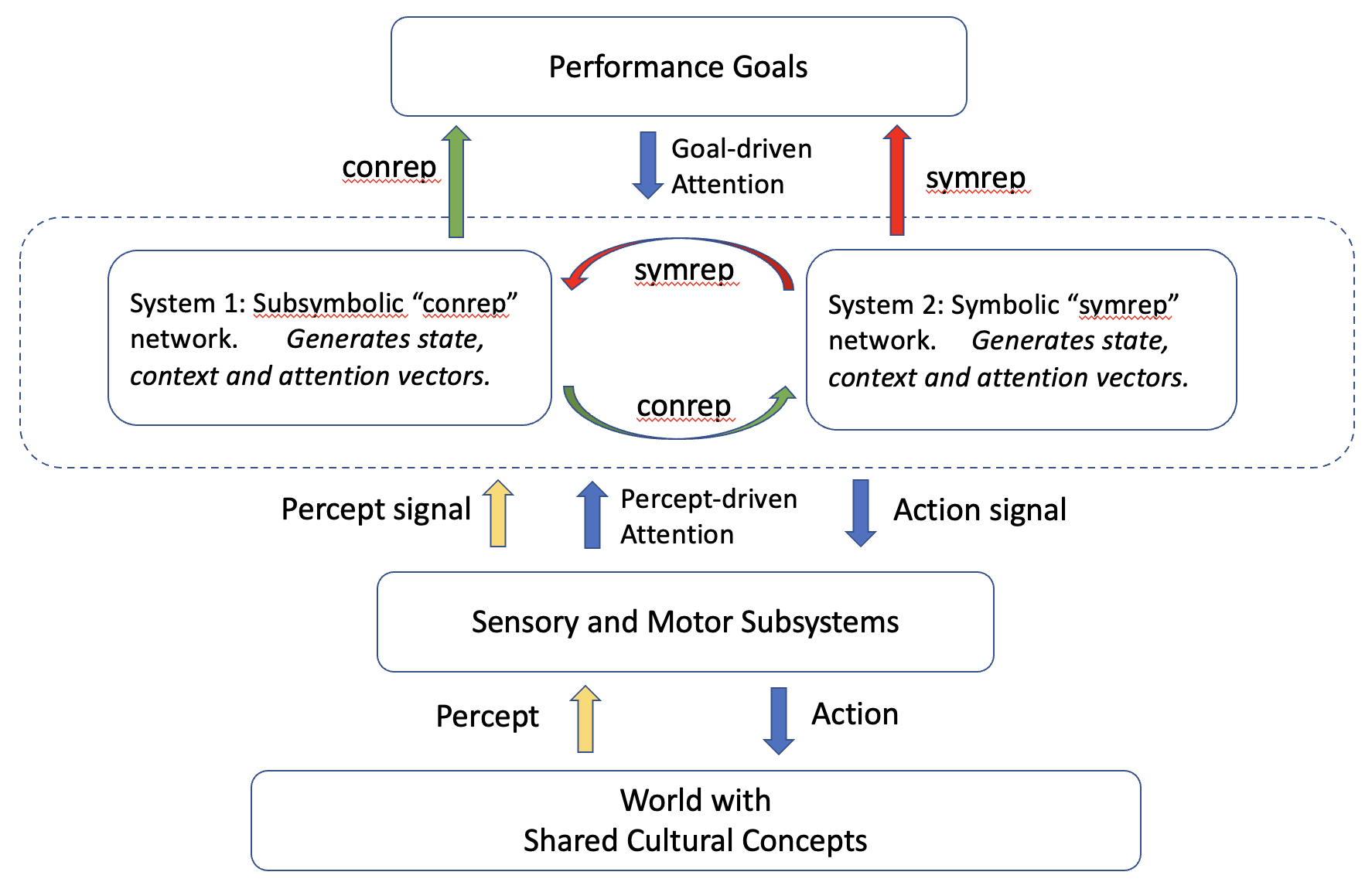}
    \caption{{\bf A Neuro-Symbolic Architecture.}  System 1 is semantically organized based on  associations/relations between representations for concepts that allows compositionality over those conreps via superposition (like the overlay of two images) using vector mapping functions.  System 2 is syntactically organized based on association/relations between representations for symbols that allows compositionality over those symrep in accord with concatenative syntactic rules (like that of logic and math) using vector mapping functions.
}
    \label{fig:symbolic-subsymbolic-arch}
\end{figure}

\vspace{0.1in}
Figure \ref{fig:symbolic-subsymbolic-arch} proposes a neuro-symbolic architecture for an intelligent agent such as a human.  It borrows from the basic diagram of an Intelligent Agent \cite{russelANDnorviq2010} 
as well as work in neuroscience 
\cite{PULVERMULLER2013},  in cognitive  psychology \cite{kahneman2011thinking}, and from work in deep neural networks \cite{Bengio2003}.
It is also inspired by recent work discussed in the neuro-symbolic literature \cite{DBLP:conf/ijcai/LambGGPAV20,DBLP:conf/iclr/MaoGKTW19}.
Here we present the four major components of the architecture, and then discuss below the more subtle points. The four components are:

\begin{itemize}
    \item \emph{Sensory and Motor Subsystems} that receive raw external percepts (images, sounds, taste, smell, touch) from the real world and provide the appropriate percept signals to the higher order System 1 and 2 components. They also receive symrep and conrep vectors from System 1 and 2, which (a) influence perceptual processing and (b) produce action sequences (motion, sound) that can affect their external world.

    \item A \emph{System 1 attractor network} that, given a percept signal, a symrep vector which contains recent context and attention information, and a goal-driven attention vector, learns to relax into an activation state we call a \emph{conrep}.  
    This conrep will represent one or more of the previously learned concepts, each at some appropriate level of abstraction, as well as relationships among these concepts.
    System 1 contains the higher-order sensory and motor functionality that is able to form an overall representation for a concept.
    System 1 is semantically organized and largely grounded in sensory/motor representations. This semantic organization has the properties that (1) two concepts with similar meanings have similar representations, and (2) certain operations over pairs of concepts can be performed via simple operations on their conrep vectors (e.g., union, summation, superposition).
    Note this System 1 network feeds back its conrep to the System 2 network, and these two networks can settle into a combined lower energy state during slow thinking. 
    
    \item A \emph{System 2 attractor network} that, given a percept signal, a conrep vector which contains recent context and attention information, and a goal-driven attention vector, learns to relax into a desired activation state we call a \emph{symrep}. The symrep will represent one or more of the previously learned symbols or some proto-symbol (to be discussed below). 
    System 2 is organized so that (1) symbols with similar appearance have the same symreps, and (2)
    symreps are organized in a way that supports compositionality in accord with concatenative syntactic rules (like that of logic and math) using vector mapping functions.   Functions over distinct concepts represented by distinct symbols can be implemented in System 2 through learned mapping functions (e.g., lookup tables for multiplication give the symrep for 63 from the inputs 7 and 9).
    
    \item A \emph{Performance Goal Subsystem} that activates relevant goals and associated attention that influences the formation of subsequent conrep and symrep based on the agent's recent perceptual input and internal state.  
    Note that performance goals (e.g., to eat) can be closely associated with internal senses (e.g. hunger).  These goals, and context and attention vectors driven by recent conrep, have significant influence on the training and relaxation of the Systems 1 and 2 attractor networks.  To some extent, agents with this architecture see what they want to see from the percepts because it helps them make sense of their world.
\end{itemize}

To illustrate in more detail the proposed architecture of Figure \ref{fig:symbolic-subsymbolic-arch}, consider how it would respond to seeing a peach versus reading the word ``peach". When this agent sees a peach, the sensory system encodes the original retinal signals or pixels into the appropriate perceptual signal and percept-driven attention for input to System 1 and 2.  System 1 relaxes into a state very similar to a previously learned state, i.e. the conrep for the concept peach. Given the percepts of this particular peach might not fit exactly the prototypical peach (i.e., this peach might have a bruise at the top), the exact relaxation state will reflect both the recalled prototypical concept, plus perceived deviations.  System 1 then automatically activates or primes activation for semantically related concepts such as fruit, sweet, and round.  Meanwhile, System 2 attempts to relax into a stable state but it cannot based on only the percept signal.  However, as System 1 forms a stable conrep, it provides input to System 2 that allows it to relax into the symrep for the word ``peach" which is  semantically related to other word symbols such as ``fruit", ``sweet", ``round", as well as phonetically related to words such as ``punch", ``pinch" and ``pooch".  This may account for why an agent is able to initially describe an object in terms of its related characteristics but not recall its name. This may also account for an agent properly classifying an object but overtly labelling it with the wrong symbol. Humans will say ``Oh yes, I knew what it was, but called it the wrong name". 

In contrast, when our agent instead sees the word ``peach" in the absence of the actual fruit, the sensory system encodes the original retinal image or pixels as an appropriate perceptual signal and percept-driven attention for input to System 1 and 2.  System 2 relaxes into a previously learned state, i.e. the symrep for the word symbol ``peach" which is semantically and phonetically related to other symbols as described above. Meanwhile, System 1 attempts to relax into a stable state but cannot based on just the percept signal.  However, as System 2 forms a stable symrep, it provides input to System 1 that allows it to relax into the conrep for the concept peach which is semantically related to other concepts such as fruit, sweet, and round. This may account for why an agent can recognize a word symbol but initially not be able to recall its meaning. In such a case, humans will tend to search for contextual clues as to where they had seen the symbol before, who had said it, or where it had been written. 

As we have described above, subsymbolic neural activity in the form of a conrep can become associated with a symbol in the form of a symrep and vice versa.  
In fact, there is evidence to suggest that an associated shared symbol is not required to do simple reasoning using a conrep.
Prior to learning the culturally shared symbol for a particular concept such as a type of fruit (apple) or a type of vegetable (peas), a child is able to identify and ask for such through gestures (hand waving, smacking lips, crying).  This suggests that humans create and use conreps for reasoning, independent of any particular culturally shared symbol.  It is interesting to ask whether humans create personal symreps as a form of proto-symbol before learning the culturally shared word symbols, and if so, why and how these symrep are used.  Children surely obtain the concept of ``mother" long before they know the word, though it is unclear whether they have only a conrep, or also an internal symrep for a proto-symbol of mother.  One possibility is that we create such symreps, based on sensory and emotional dimensions (e.g., apples are sweet and make us  happy) and learned associations about them (e.g., apples often come from mother).  


Although learned associations, such as ``apples often come from mother", might be encoded at the conrep level, there might be a significant energy-related advantage to using symreps.  If activating a conrep requires significantly greater neural activity than activating its associated symrep, then precious energy might be saved by instead activating and reasoning in terms of symreps.  Why and when might this be the case?  If conreps provide a full multisensory representation of an object and therefore generate brain-wide neural activity across many sensory-motor cortical regions (see Figure \ref{fig:distributedEncodings}), they will be high dimensional vectors of activity that burn significant energy.  In contrast, if symreps need only encode the identity and not all of the perceptual features of an object, they can be much lower dimension and require correspondingly less energy (e.g., a bit vector representation needs only $\log_2 N$ dimensions to provide a symrep for a vocabulary of $N$ symbols).  One can imagine reasoning systems that operate efficiently in terms of symreps (e.g., making a sequence of inferences based on memory retrievals and learned associations among symreps), then activate corresponding conreps only for the reasoning steps that require sensory-motor representations.  An intermediate hypothesis is that the brain simply activates only a portion of the conrep that is needed for the inference at hand, perhaps through an attention mechanism that selects which portion to activate.


A neuro-symbolic architecture such as the one we propose above must be able to perform a number of important functions at the core of  an intelligent agent. These include: communicating, reasoning, and learning. Each of these will be discussed in the following section.

\section{Discussion}

\subsection{Symbols for External and Internal Communication}

Symbols are clearly the fundamental tools used for external communication by an agent to another agent.  Furthermore, the rules and constructs of language employ symbols to make communications as efficient and effective as possible. These rules and constructs can be considered symbols in their own right, for example we understand that ``A, B, or C" means ``either A or B or C".  However, others come with ambiguity, for example consider the phrase ``time flies like sand".

Less evident is the use of symbols to refer internally to complex multi-sensory abstractions (conrep) that assist with inference and reasoning.  We propose that symbols embodied by their symrep are used literally as a ``language of thought" in the brain and could similarly be used in a system of ANNs to explain a decision and to provide important constraints to otherwise subsymbolic (conrep based) responses.

Overt (oral, written) symbol manipulation is typically a conscious System 2 activity. However, we argue that symreps activated by System 2 can affect unconscious System 1 activity, constraining the relaxation of the conrep network to a solution that satisfies all percepts.
Symbols (particularly words) are used to organize long-term memory with respect to spelling, phonetics, syntax (as part of a phrase) and semantics.  In these ways, symrep constrain learning and inference beyond the fundamental semantics of conrep that would tend to associate concepts based on factors such an objects size, colour, and shape.  Culturally shared symbols (e.g., words) are a mechanism for each of us individuals to be taught the most relevant general categories and concepts learned by those who came before us.

\subsection{Reasoning using Representations of Symbols and Concepts}

Formal symbolic reasoning in humans is considered a conscious use of a formal language such as logic or algebra to execute production rules or mathematics on symbols to infer a result.  Formal reasoning leads the agent from some given symbolic state through a series of deductive steps to a new result state.  But many types of reasoning (e.g., how hard to press the brake pedal to avoid hitting the pedestrian) seem to operate at a subsymbolic level.

One of the key open questions in AI is how can we design artificial neural networks that integrate symbolic knowledge such as Founder(Amazon)=JeffBezos and Sum(8,9)=17 with subsymbolic reasoning.  Similarly, one of the key open questions in cognitive neuroscience is how the human brain manages to perform symbolic reasoning despite the fact that the brain is implemented by neurons using distributed representations.

We propose that the brain reasons using a combination of symreps (distributed vectors of neural activity that represent symbols) and conreps (distributed vectors of neural activity that represent concepts referred to by symreps), and that future AI systems will benefit from the same dual representations.  Our conjecture involves the following assumptions which we will discuss further below:

\begin{itemize}


\item A symbol's symrep acts as an {\em index} into the conrep of the concept referred to by the symbol.  Similarly, the conrep can act as an index into the synrep of its corresponding symbol.  Here when we say that ``A is an index into B" we mean simply that the distributed neural activity A can cause neural activity B.  We assume there is a neural mechanism that implements both (1) the activation of the corresponding symrep (conrep) given the conrep (symrep), and (2) the completion of a multimodal conrep given just a partial perception of it (e.g., activate the taste of a peach when we see one). \cite{DBLP:conf/flairs/IqbalS16} provides an example of such a trainable ANN system, and presents evidence showing its ability to perform both of these functions.

\item Reasoning, which is a matter of producing future neural activity from current neural activity, can take place either by manipulating symreps or conreps, or both.
Within our proposed architecture, rational decision making, math, and logic take place within System 2 that is composed of some form of recurrent attractor network that can rewrite sequences such as (2+3)*4 with 5*4, and 5*4 with 20. 

\item One advantage of performing reasoning in parallel at both the level of symreps and conreps is that these two processes can guide and constrain one another, because of the brain's ability to index back and forth between symreps and conreps.

\item Another advantage of the dual system is that by replaying the sequence of steps taken by System 1 to relax into a stable state, System 2 can be used to explain aspects of that sequence. We suggest this is what happens when one is asked to explain how they reached an important decision or how they hit the ball so well. 


\end{itemize}

\begin{figure}[ht]
    \centering
    \includegraphics[width=0.99\textwidth]{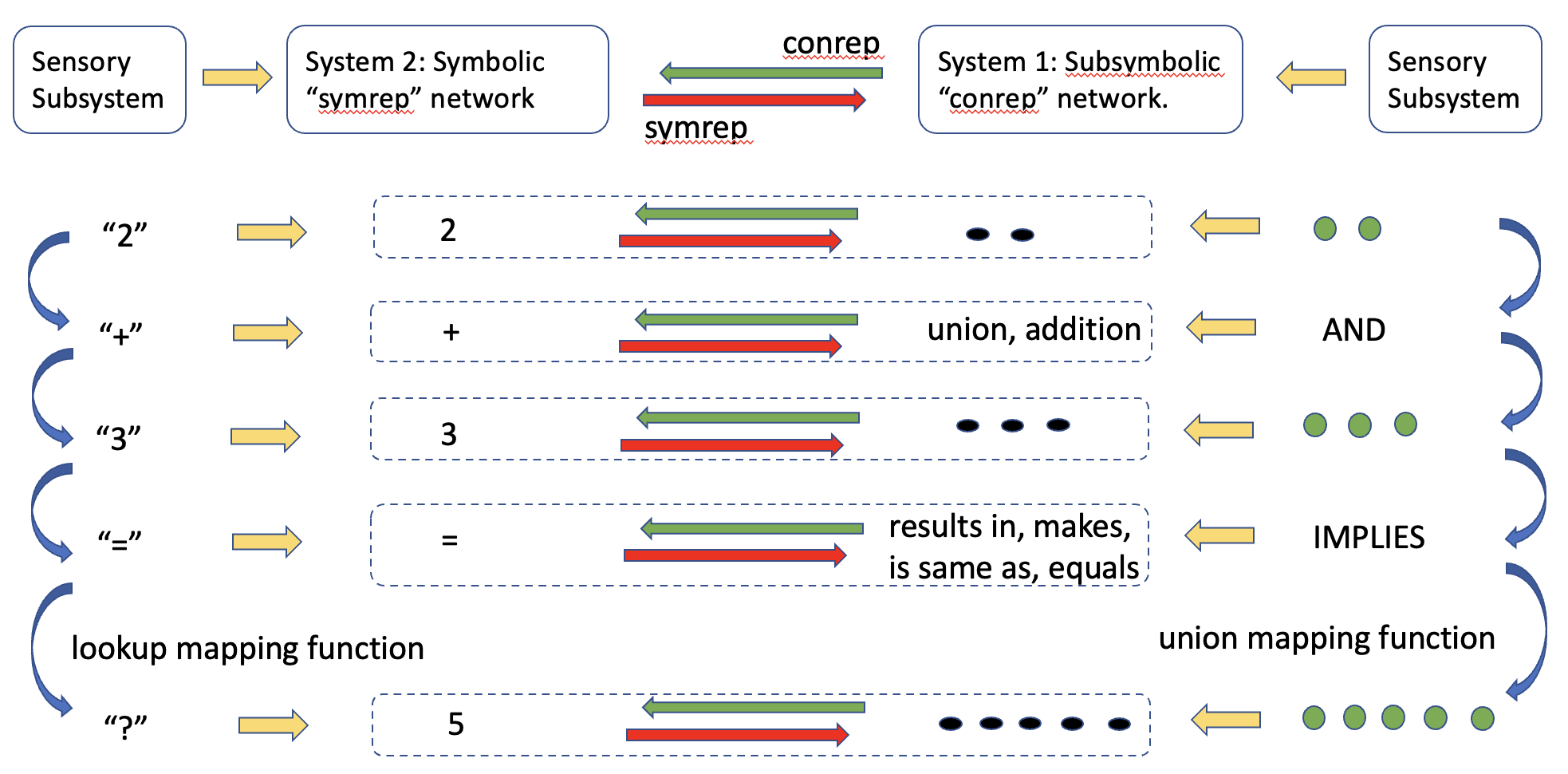} 
    \caption{{\bf Reasoning may occur using neural representations of symbols (symreps, left), or neural representations of concepts (cconreps, right), or both.}  See text for explanation. }
    \label{fig:reasoning}
\end{figure}

Figure \ref{fig:reasoning} illustrates this conjecture for a simple case of reasoning about the question of how many marbles one has if they add three new marbles to the two they currently have. 
The right side of the figure depicts the conrep System 1 network reasoning path. We assume the conrep for ``two marbles" includes neural activity in a part of the brain that can represent and manipulate small numbers (or sets) of items in a fashion that can be easily perceived and manipulated. Using this conrep, we conjecture, the brain can combine (superimpose, create the union of) its representations of ``two marbles" and ``three marbles" to create the conrep of ``five marbles", answering the question.  The left side of Figure \ref{fig:reasoning} depicts the disjoint reasoning path of the symrep System 2 network.  Here, we assume the perceived two marbles activates the symrep for the symbol ``2", and the perceived new three marbles activates the symrep for ``3".  Given these symreps, the reasoning step is now a matter of recalling from memory that ``2+3" equals ``5."  This also answers the question and supports the reasoning pathway of of System 1.

Notice the difference in the conrep and symrep reasoning paths.  The conrep System 1 path relies on the ability of the brain to reason non-symbolically by operating directly on the conreps; in this case, operating on the conreps of two small sets to get the conrep of their union.  In contrast, the System 2 symrep leval path relies instead on a memorized addition table, and the ability to index easily into this table using the symreps for ``2" and ``3", as well as the ability to activate the correct table given the symrep for ``addition".  
Notice also that there is no reason that these two reasoning paths cannot occur simultaneously in the brain, and no reason that these two paths cannot cross-communicate with one another -- in fact, our proposed architecture assumes both.
For example, each step of the System 1 pathway on the right side of Figure  \ref{fig:reasoning} can index into (i.e., activate) the symrep obtained by the System 2 pathway on the left.  Similarly, the System 2 reasoning path can activate conreps on the reasoning path of System 1.  If these two pathways do not agree, one can imagine a process that detects and repairs such a disagreement.  Finally, note that if one of these two processes outpaces the other, obtaining an result more quickly, then it can serve to prime the neural activity for the not-yet-reached result of the other path.  Even though we still have one head, in this sense two paths are better than one. 

Finally, we conjecture that in some cases, we simply have no symbolic representations, and rely solely on subsymbolic processes.  As an example, imagine answering the question of what it would taste like if you put peanut butter on a peach.  You can think of the answer, and almost taste it, but cannot put your answer into words.  In yet other cases, we have {\em only} symbolic representations, and must rely solely on neural processes that manipulate the symreps.  As an example, consider the question of what is 1,539 times 20?  Here we can answer by reasoning symbolically, but our number sense at the level of conceptual representations does not include representations for such large numbers at full resolution. In fact, neuroscience and behavioral studies (see \cite{Volk2020} for an overview) indicate that humans have an automatic ability to perceive and operate on small quantities (we automatically perceive that the customer has three lemons at the checkout counter, and not two), but use different methods and different brain regions for larger numbers (we must count so see whether they have 12 pieces of corn).  Interestingly, \cite{DBLP:conf/nesy/SilverG21} found their LSTM based ANN for arithmetic developed a better model for arithmetic when forced to encode input symbols (e.g., 2+3) as thermometer style conreps at the output (see the discussion of this system below).


\subsection{Symbols as a Source of Inductive Bias when Learning}

We suggest that symbols are not only beneficial to reasoning, but that they are also instrumental in learning new concepts.
Specifically, we propose that symbols act as important constraints (inductive biases) on the formation of new representation for concepts.
Many symbols are actually categories themselves (e.g., fruit, politics) and are a way of culturally communicating which categories/generalizations our predecessors have found useful to know and think about. These, of course, change over time along with cultural norms.


The correct recognition of a symbol acts as a source of inductive bias for learning more complex reasoning patterns that make use of that symbol.
Consider a first grader learning numbers and basic addition.  Children learn to represent symbols such as the digit ``3" and word ``three" by a symrep vector of neural activity. 
The symrep for ``3" becomes closely associated with the conrep for the numeric concept three that is associated with other sensory/motor percept components such as seeing three dots, wiggling three fingers, hearing three knocks, feeling three touches, etc. 
When a first grader is learning to ``add 2 apples to 3 apples", the student must first correctly activate the conrep for the concept $\langle$ three $\rangle$, else the reasoning sequence will result in an error.
Therefore, prior learning of the symrep for a symbol, and its mapping to its associated conrep, increases the probability of correctly using the conrep as part of a larger reasoning activity.
The young student learns the correct reasoning sequence for ``add 2 apples to 3 apples" in part because activating the correct symrep for the digit ``3" will ensure the student activates the correct conrep for the concept three that forms part of the input for the next reasoning step. 

We have modeled this recently in a study reported in \cite{DBLP:conf/nesy/SilverG21}.
There, we train recurrent LSTM neural networks to perform arithmetic operations (+,-,×,/) on sequences of images of noisy handwritten digits and operators. 
We desire a learning system that is able to learn: (1) the quantity that each digit represents, (2) the ordinal relationship between one digit and another, also known as successorship, and (3) the function of the operator.  (1) and (2) need to be contained in the conrep for a digit, and (3) is what needs to be contained in the conrep for the operator.
Human learners are able to develop conceptual representations (conreps) for both the digits and the operators that capture the appropriate semantics. The output training labels provided to our neural network models must represent the same semantics. 
Although a one-hot vector representation, such as
“000001000” for the digit “5” is unabiguous, it lacks the representational ability to convey the first two semantics concerning digits: it does not convey quantity and it does not represent an ordering over the digits. 
We can alternatively use a unary or thermometer encoding that is a binary vector with as many elements sequentially set to one as the integer being represented (e.g., the number 5 is encoded as “000011111”). This representation conveys the semantics of (1) and (2) concerning digits and can be used to develop an appropriate conrep.  In fact, by training our system to output such thermometer encodings for each noisy input image, as well as for the calculated output product, we are essentially training it to a particular conrep.

\begin{figure}
	\centering
	\includegraphics[width=4.8in]{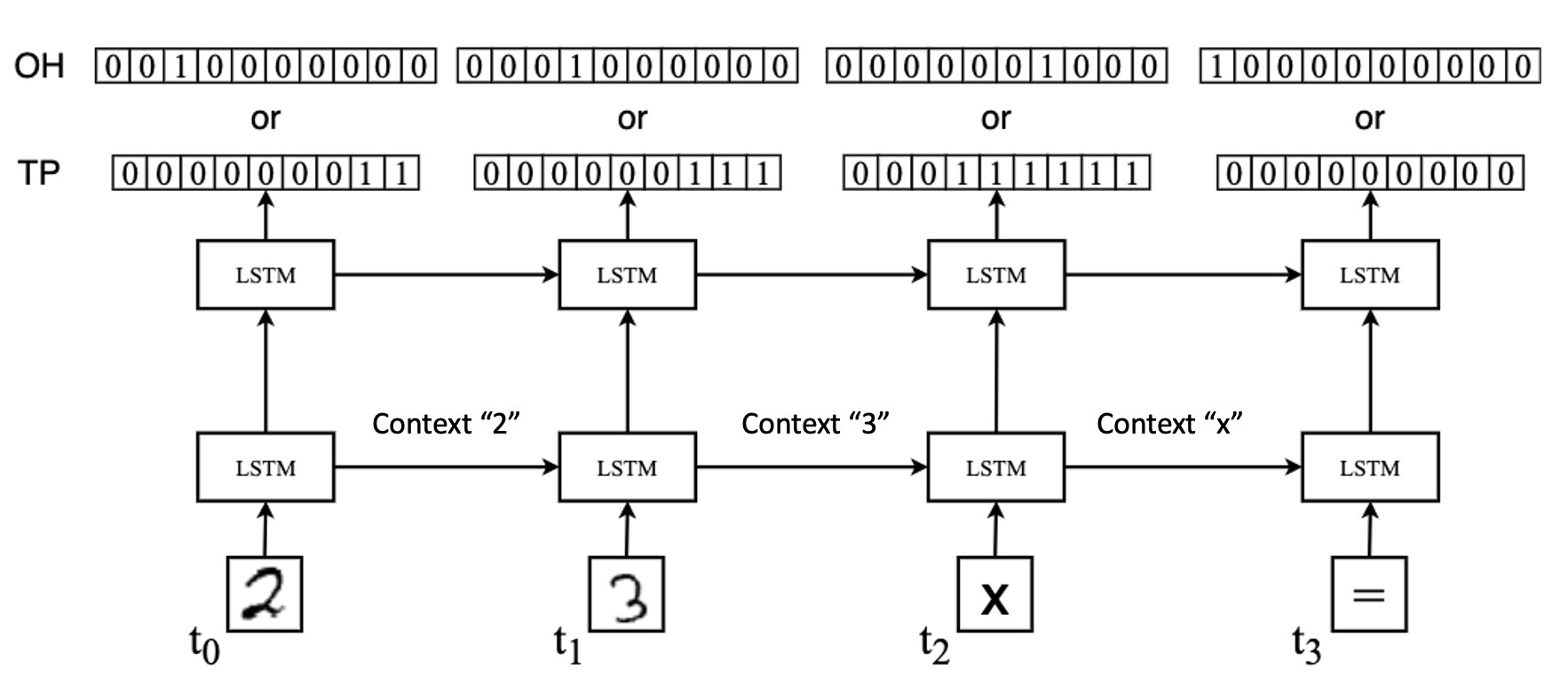}
	\caption{A sequential model learning to perform arithmetic (in this case, 2 x 3) on images of handwritten digits in the presence of symbolic output representations. The symbolic representations are provided on the outputs of the first two intermediate steps using either one-hot (OH) or thermometer (TP) encoding.}
	\label{fig:sequential-model-with-classes}
\end{figure}

Figure \ref{fig:sequential-model-with-classes} depicts the input and output layers and the architecture of the recurrent network on each sequence of four time steps.
The recurrent network inputs a 28 x 28 grey scale image at each step (i.e., the input symbols).
The network can use its output nodes for two different tasks: (a) to output its encoding of the noisy input digit image, and (b) to output its calculated the result of the arithmetic operation.
We trained this network in two ways: in the presence and absence of output conceptual representations for the noisy digits; that is, the network either learns just the arithmetic operation task (b) or both (a) the digit conrep classification task and (b). Furthermore, we also trained it using either one-hot or thermometer encodings of numerical digits.
The output layer consists of 10 dimensional representation of the numeric values, represented either as the output as either a one-hot vector or a thermometer vector.  
Our experiments show that the models trained to output an encoding of each noisy handwritten input digit are more effective at performing arithmetic than the models trained without this additional task. 
Furthermore, we show when the recurrent network models are trained to produce thermometer encodings of their outputs, they discover more accurate algorithms for all arithmetic operations for both seen and previously unseen combinations of digits. 

%

\begin{figure}[ht]
	\centering
	\includegraphics[width=0.8\textwidth]{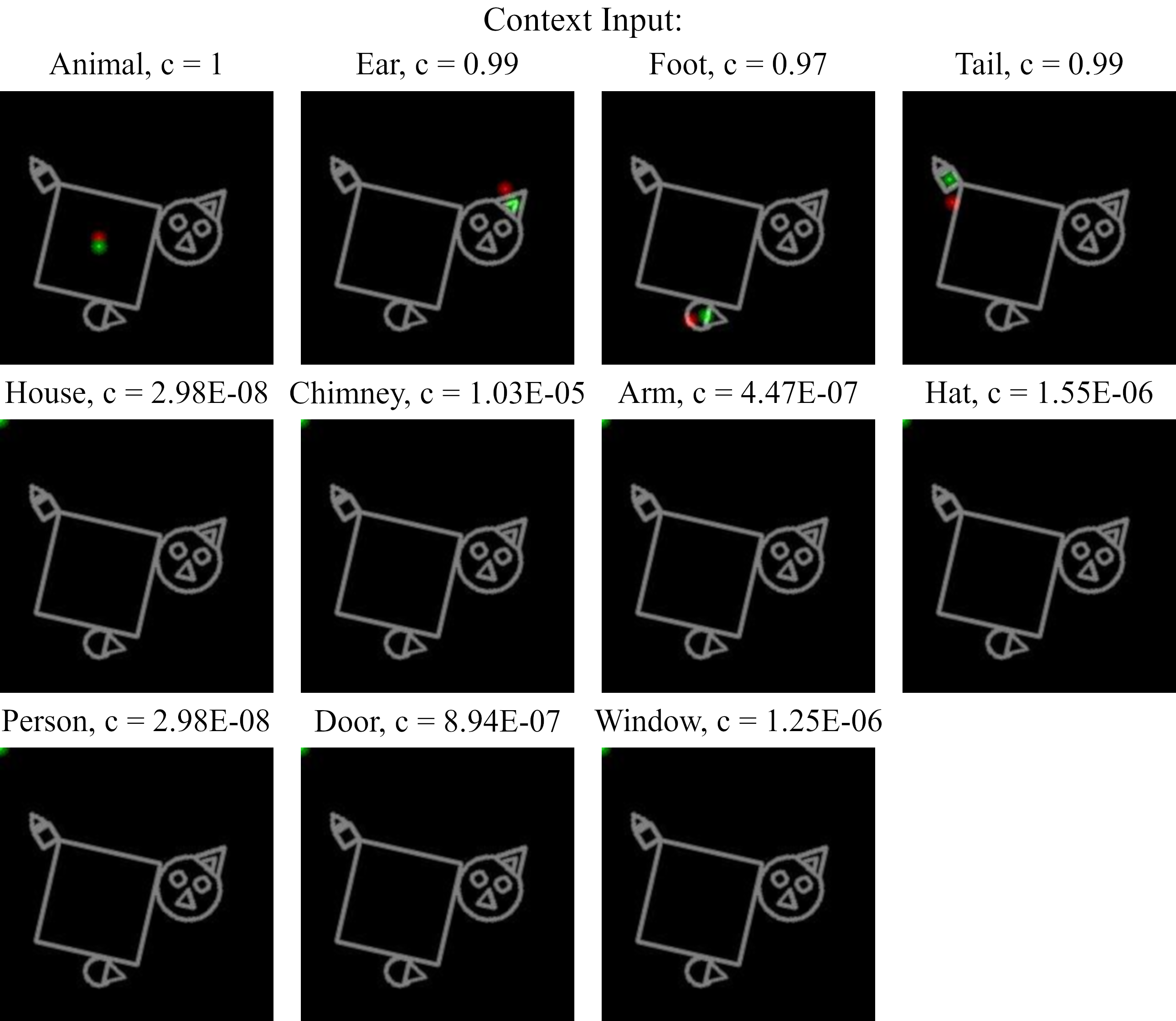}
	\caption[A correctly classified instance with all features also classified correctly]{A correctly classified instance with all features also classified correctly. The value of c is the predicted probability of the class. Red dots are predicted locations, green dots are labelled locations.}
	\label{fig:COR_EXPL1}
\end{figure}

In separate work, we have also shown that it is possible to learn to classify objects in images more accurately and to explain misclassifications by developing deep neural networks that learn to locate and classify both objects and their features in the images \cite{Hiltz_thesis_2020}. 
This requires using Multiple Task Learning (MTL) networks to locate and clasify objects (e.g. animal, person, house) as well as their features (e.g. head, feet, ears, windows, doors, chimney).
This study shows that the best MTL ANN architecture for this task is context sensitive MTL (csMTL), which uses context inputs to select which task the network’s outputs are classifying and localizing \cite{DBLP:journals/ml/SilverPC08}. Feature classification in parallel with object classification results in a significant boost in F1 score for object classification. Examples demonstrate how the explanations given by the feature classifications or localizations, when in error, lead to the misclassification of objects.
Figure \ref{fig:COR_EXPL1} shows pictures of an animal with true and predicted localization coordinates. Labelled locations for features and objects are indicated by the green dots and the red dots are used for predicted locations. This example contains all three possible labelled features that an animal can possess and the classifier correctly identified and localized each of them. Furthermore, only the first four classes of objects output by the csMTL network (animal, ear, foot, tail) produced a high probability of being present. Activation of the two remaining objects and five remaining feature context inputs produce near-zero valued outputs for both presence and location. Thus, this classification is correct and feature predictions are consistent with this object class, justifying the classification.

More generally, symbols help to topologically, semantically and hierarchically organize the things we learn over a lifetime in an associative manner.  The symrep for the word symbols ``ears", ``eyes", ``nose" and ``mouth" have relationships between one another that are learned because of the subsymbolic visual relationships between the conrep for each of the same concepts.  Therefore, symbols are useful for creating concepts in memory and for searching for concepts in memory.  For example: you can retrieve a person’s name based, in part, on a description of their face (long nose, blue eyes, small mouth).

\section{Conclusions}

\subsection{Summary}

Although it may be possible to build intelligent agent architectures that do not use external symbols, we suggest here that if an agent has the ability to associate internal concepts (vectors of neural activity) with external symbols, and use those symbols to communicate, then a number of wonderful things can happen.
Symbols provide the means by which knowledge within the nervous system of one agent can be used to inform the nervous system of another agent, without that second agent expending energy in the real world or risking its life or limb.  This includes the use of language by one agent to inform another agent about the current state of its world, but even more importantly, the use of language to express more general information about which concepts, or generalizations, are important, and which problem solving methods work.  

More importantly, we conjecture that internal agent “self-communication” using the same symbol representations meant for agent-to-agent communication, has become key to human intelligence because: (1) it provides a second, more abstract level of internal representation and reasoning which can occur in parallel with subsymbolic reasoning, and (2) it places an additional constraint on learning where prior learning of accurate symbol classifications models act as an inductive bias for learning new symbols and concepts.   
Shared symbols allow us to explain and justify, internally as well as externally, our decisions and actions.  And what we learn is shaped and constrained by the “lexicon” of what we recognize as symbols. 

In this paper we have introduced terminology for discussing the representations and processing of symbols and their associated concepts in neuro-symbolic systems.  Furthermore, we have proposed a 
plausible architecture for intelligent agents that uses and combines subsymbolic representations for symbols and concepts for learning and reasoning. 
Key features of our proposed architecture include:

\begin{itemize}
\item
Both symbols and the concepts they refer to are represented by distributed vectors of neural activation, which we call symreps and conreps respectively.  For example, the human brain uses one vector of neural activity to represent the appearance of the letter string symbol “peach” (the symrep), and another vector of neural activity to represent the meaning of this symbol (the conrep) which is associated with or has as a modal part the symrep for the symbol “peach”).  Brain imaging studies reveal that conreps are surprisingly similar across humans, to the degree that machine learning classifiers trained on brain image data (i.e., a facimile for their conreps) of person A can be used to decode the concept about which person B is thinking, without training on the brain image data for person B.

\item
Conreps and symreps cross index one another, in the sense that the neural activity for a symrep can be used to generate the neural activity for the corresponding conrep, and vice versa.

\item
This ability of conreps and symreps to activate one another yields a kind of introspection, in which the architecture can approximate using symbols its subsymbolic representations and reasoning, allowing it to communicate externally about its internal mental states.

\item
Decision making and problem solving always takes place by propagating distributed neural activity, but that neural activity can include the processing of conreps, or symreps, or both in parallel, depending on the specific problem.  This leads to multiple, parallel processing streams working to make the same decision, and to the ability of the architecture to synchronize these parallel processing streams through the cross indexing of conreps and symreps.

\item
The multiple representations and multiple processing streams supported by conreps and symreps align to some degree with the System 1 and System 2 model of human cognition advocated by Daniel Kahneman in his book “Thinking Fast and Slow" \cite{kahneman2011thinking}.  Fast thinking in his System 1 corresponds to neural processing of conrep activations, whereas the slower thinking of System 2 corresponds to sequential reasoning steps involving symreps, conreps or both.

\item
External symbolic communication plays a key role in the agent’s learning, including providing a culturally shared inductive bias for generalization from sets of examples. For instance, the mere existence of the word “fruit” suggests that the set of fruits is a useful general category, whereas many possible sets have no corresponding word such as the collection of $n$ random products from a hardware store.  Similarly, verbs such as “grab” describe a large set of actions that can usefully be grouped together for general reasoning and problem solving.
\end{itemize}

Thus, in our proposed architecture symbols are not the fundamental building blocks of all thought.  But they are characterizations of neural activity (conreps) that (1) enable us to explain our subsymbolic thinking to ourselves and others, and (2) act as constraints on inference and learning.  Symbols explain and aid our thinking, but are not the fundamental basis for our reasoning.

\subsection{Future Work}

We suggest that evolutionary pressure could have led to the proposed neuro-symbolic architecture, as the ability to communicate among agents will yield great advantages to fitness of the species. Furthermore, we suggest there might be other evolutionary pressures such as minimizing energy consumption in the brain that could lead to brains that use, whenever possible, lower dimensional symreps that require much less neural activation than the more complete but more energy-consuming conreps. These are interesting areas of research for evolutionary biology and neuroscience.

Of course this discussion raises many other questions for future work.
Context plays an important role in choosing the actions that are taken for any given percept. How is context represented and used?
Attention to the most relevant aspect of a percept is required by an agent in order to survive in the world. 
What are the sources of attention and and how does attention interact systematically with both conreps and symreps?  

Continual lifelong learning of symbols and concepts is a must for an intelligent agent.
How do we consolidate new symbols (symrep) and concepts (conrep) into an agent's long-term memory?
Can we empirically study ANNs, especially transformer-based architectures of LLMs such as GPT3, to gain more insight into how they might integrate symbolic and subsymbolic representations and reasoning?  

Imagination includes the ability to think of concepts we have never seen. How do we imagine new solutions to challenging complex problems? Is this a process that depends on the generation of new symbolic representation (symrep) that resonates with new conceptual representations (conrep)?  

We look forward to an increased understanding of neuro-symbolic systems over the coming years of progress across the fields of neuroscience, cognitive science and artificial intelligence.

\section{Acknowledgements}

We thank Andrew Mcintyre and Robert Mercer for useful comments on earlier drafts of this paper.  We acknowledge the support of AFOSR for this research, under grant FA95501710218 and to the Harrison McCain Foundation for Visitorship funding.

\bibliographystyle{tfnlm}
\bibliography{Chapter-Silver-Mitchell/reference}

\begin{thebibliography}{10}
\providecommand{\url}[1]{\normalfont{#1}}
\providecommand{\urlprefix}{Available from: }

\bibitem{jung1968}
Jung~CG. Man and his symbols. Vol. 5183. Dell; 1968.

\bibitem{CSPeirce_Wikipedia}
Wikipedia. Charles sanders peirce ; 2022.
  \urlprefix\url{https://en.wikipedia.org/wiki/Charles_Sanders_Peirce}.

\bibitem{rumelhart1988}
Rumelhart~DE, McClelland~JL, Group~PR, et~al. Parallel distributed processing.
  Vol.~1. IEEE New York; 1988.

\bibitem{FELDMAN1982}
Feldman~J, Ballard~D. Connectionist models and their properties. Cognitive
  Science. 1982;\hspace{0pt}6(3):205--254.
  \urlprefix\url{https://www.sciencedirect.com/science/article/pii/S0364021382800013}.

\bibitem{Mikolov2013}
Mikolov~T, Chen~K, Corrado~GS, et~al. Efficient estimation of word
  representations in vector space. In: International Conference on Learning
  Representations; 2013.

\bibitem{Mitchell2008}
Mitchell~TM, Shinkareva~SV, Carlson~A, et~al. Predicting human brain activity
  associated with the meanings of nouns. Science. 2008;\hspace{0pt}320:1191 --
  1195.

\bibitem{Sudre2012}
Sudre~GP, Pomerleau~DA, Palatucci~M, et~al. Tracking neural coding of
  perceptual and semantic features of concrete nouns. NeuroImage.
  2012;\hspace{0pt}62:451--463.

\bibitem{Mitchell2004}
Mitchell~TM, Hutchinson~RA, Niculescu~RS, et~al. Learning to decode cognitive
  states from brain images. Machine Learning. 2004;\hspace{0pt}57:145--175.

\bibitem{shinkareva2008}
Shinkareva~S, Mason~R, Malave~V, et~al. Using fmri brain activation to identify
  cognitive states associated with perception of tools and dwellings. PLOS ONE
  3(1): e1394. 2008;\hspace{0pt}.

\bibitem{haxby2001}
Haxby~JV, Gobbini~MI, Furey~ML, et~al. Distributed and overlapping
  representations of faces and objects in ventral temporal cortex. Science.
  2001;\hspace{0pt}293(5539):2425--2430.

\bibitem{Just2012}
Buchweitz~A, Shinkareva~SV, Mason~RA, et~al. Identifying bilingual semantic
  neural representations across languages. Brain and Language.
  2012;\hspace{0pt}120(3):282--289.
  \urlprefix\url{https://www.sciencedirect.com/science/article/pii/S0093934X11001568}.

\bibitem{PULVERMULLER2013}
Pulvermüller~F. How neurons make meaning: brain mechanisms for embodied and
  abstract-symbolic semantics. Trends in Cognitive Sciences.
  2013;\hspace{0pt}17(9):458--470.
  \urlprefix\url{https://www.sciencedirect.com/science/article/pii/S1364661313001228}.

\bibitem{Sassenhagen2020}
Sassenhagen~J, Fiebach~CJ. {Traces of Meaning Itself: Encoding Distributional
  Word Vectors in Brain Activity}. Neurobiology of Language. 2020
  03;\hspace{0pt}1(1):54--76.
  \urlprefix\url{https://doi.org/10.1162/nol\_a\_00003}.

\bibitem{FabreThorpe2001ALT}
Fabre-Thorpe~M, Delorme~A, Marlot~C, et~al. A limit to the speed of processing
  in ultra-rapid visual categorization of novel natural scenes. Journal of
  Cognitive Neuroscience. 2001;\hspace{0pt}13:171--180.

\bibitem{Jigo2020}
Jigo~M, Carrasco~M. Differential impact of exogenous and endogenous attention
  on the contrast sensitivity function across eccentricity. Journal of Vision.
  2020;\hspace{0pt}20.

\bibitem{Dugu2018}
Dugu{\'e}~L, Merriam~EP, Heeger~DJ, et~al. Differential impact of endogenous
  and exogenous attention on activity in human visual cortex. Scientific
  Reports. 2018;\hspace{0pt}10.

\bibitem{Jigo2021}
Jigo~M, Heeger~DJ, Carrasco~M. An image-computable model of how endogenous and
  exogenous attention differentially alter visual perception. Proceedings of
  the National Academy of Sciences. 2021;\hspace{0pt}118.

\bibitem{kahneman2011thinking}
Kahneman~D. Thinking, fast and slow. New York: Farrar, Straus and Giroux; 2011.
  \urlprefix\url{https://www.amazon.de/Thinking-Fast-Slow-Daniel-Kahneman/dp/0374275637/ref=wl_it_dp_o_pdT1_nS_nC?ie=UTF8&colid=151193SNGKJT9&coliid=I3OCESLZCVDFL7}.

\bibitem{Stanovich2000}
Stanovich~KE, West~RF. Individual differences in reasoning: Implications for
  the rationality debate? Behavioral and Brain Sciences.
  2000;\hspace{0pt}23(5):645--665.

\bibitem{Kaas2013}
Kaas~J. The evolution of brains from early mammals to humans. Wiley
  interdisciplinary reviews Cognitive science. 2013 03;\hspace{0pt}4:33--45.

\bibitem{bommasani2021}
Bommasani~R, Hudson~DA, Adeli~E, et~al. On the opportunities and risks of
  foundation models. arXiv preprint arXiv:210807258. 2021;\hspace{0pt}.

\bibitem{chowdhery2022}
Chowdhery~A, Narang~S, Devlin~J, et~al. Palm: Scaling language modeling with
  pathways. 2022. 2022
  April;\hspace{0pt}\urlprefix\url{https://storage.googleapis.com/pathways-language-model/PaLM-paper.pdf}.

\bibitem{devlin2018}
Devlin~J, Chang~MW, Lee~K, et~al. Bert: Pre-training of deep bidirectional
  transformers for language understanding. arXiv preprint arXiv:181004805.
  2018;\hspace{0pt}.

\bibitem{GPT3_2020}
Brown~T, Mann~B, Ryder~N, et~al. Language models are few-shot learners. In:
  Larochelle~H, Ranzato~M, Hadsell~R, et~al., editors. Advances in Neural
  Information Processing Systems; Vol.~33. Curran Associates, Inc.; 2020. p.
  1877--1901.
  \urlprefix\url{https://proceedings.neurips.cc/paper/2020/file/1457c0d6bfcb4967418bfb8ac142f64a-Paper.pdf}.

\bibitem{Vaswani2017}
Vaswani~A, Shazeer~N, Parmar~N, et~al. Attention is all you need. In: Guyon~I,
  Luxburg~UV, Bengio~S, et~al., editors. Advances in Neural Information
  Processing Systems; Vol.~30. Curran Associates, Inc.; 2017.
  \urlprefix\url{https://proceedings.neurips.cc/paper/2017/file/3f5ee243547dee91fbd053c1c4a845aa-Paper.pdf}.

\bibitem{Mikolov_Word2Vec_2013}
Mikolov~T, Chen~K, Corrado~G, et~al. Efficient estimation of word
  representations in vector space. CoRR. 2013;\hspace{0pt}abs/1301.3781.
  \urlprefix\url{http://dblp.uni-trier.de/db/journals/corr/corr1301.html#abs-1301-3781}.

\bibitem{pennington2014glove}
Pennington~J, Socher~R, Manning~CD. Glove: Global vectors for word
  representation. In: EMNLP; Vol.~14; 2014. p. 1532--1543.

\bibitem{Jat2019}
Jat~S, Tang~H, Talukdar~PP, et~al. Relating simple sentence representations in
  deep neural networks and the brain. ArXiv. 2019;\hspace{0pt}abs/1906.11861.

\bibitem{DBLP:conf/flairs/IqbalS16}
Iqbal~MS, Silver~DL. A scalable unsupervised deep multimodal learning system.
  In: Markov~Z, Russell~I, editors. Proceedings of the Twenty-Ninth
  International Florida Artificial Intelligence Research Society Conference,
  {FLAIRS} 2016, Key Largo, Florida, USA, May 16-18, 2016. {AAAI} Press; 2016.
  p. 50--55.
  \urlprefix\url{http://www.aaai.org/ocs/index.php/FLAIRS/FLAIRS16/paper/view/12928}.

\bibitem{Volk2020}
Volk~JE, Parhami~B. Number representation and arithmetic in the human brain.
  2020 11th IEEE Annual Information Technology, Electronics and Mobile
  Communication Conference (IEMCON). 2020;\hspace{0pt}:0712--0717.

\bibitem{russelANDnorviq2010}
Russell~S, Norvig~P. Artificial intelligence: A modern approach. 3rd ed.
  Prentice Hall; 2010.

\bibitem{Bengio2003}
Bengio~Y, Ducharme~R, Vincent~P, et~al. A neural probabilistic language model.
  J Mach Learn Res. 2003 mar;\hspace{0pt}3(null):1137–1155.

\bibitem{DBLP:conf/ijcai/LambGGPAV20}
Lamb~LC, d'Avila Garcez~AS, Gori~M, et~al. Graph neural networks meet
  neural-symbolic computing: {A} survey and perspective. In: Bessiere~C,
  editor. Proceedings of the Twenty-Ninth International Joint Conference on
  Artificial Intelligence, {IJCAI} 2020. ijcai.org; 2020. p. 4877--4884.
  \urlprefix\url{https://doi.org/10.24963/ijcai.2020/679}.

\bibitem{DBLP:conf/iclr/MaoGKTW19}
Mao~J, Gan~C, Kohli~P, et~al. The neuro-symbolic concept learner: Interpreting
  scenes, words, and sentences from natural supervision. In: 7th International
  Conference on Learning Representations, {ICLR} 2019, New Orleans, LA, USA,
  May 6-9, 2019. OpenReview.net; 2019.
  \urlprefix\url{https://openreview.net/forum?id=rJgMlhRctm}.

\bibitem{DBLP:conf/nesy/SilverG21}
Silver~DL, Galila~A. Learning arithmetic from handwritten images with the aid
  of symbols. In: d'Avila Garcez~AS, Jim{\'{e}}nez{-}Ruiz~E, editors.
  Proceedings of the 15th International Workshop on Neural-Symbolic Learning
  and Reasoning as part of the 1st International Joint Conference on Learning
  {\&} Reasoning {(IJCLR} 2021), Virtual conference, October 25-27, 2021;
  ({CEUR} Workshop Proceedings; Vol. 2986). CEUR-WS.org; 2021. p. 154--164.
  \urlprefix\url{http://ceur-ws.org/Vol-2986/paper12.pdf}.

\bibitem{Hiltz_thesis_2020}
Hiltz~J. Explainable ai for image classification via feature prediction and
  localization ; 2020. Honours Thesis, Acadia University;
  \urlprefix\url{https://scholar.acadiau.ca/islandora/object/theses:3403}.

\bibitem{DBLP:journals/ml/SilverPC08}
Silver~DL, Poirier~R, Currie~D. Inductive transfer with context-sensitive
  neural networks. Mach Learn. 2008;\hspace{0pt}73(3):313--336.
  \urlprefix\url{https://doi.org/10.1007/s10994-008-5088-0}.

\end{thebibliography}

\end{document}